\documentclass{article}

\usepackage[nonatbib,preprint]{nips_2018}

\usepackage{microtype}
\usepackage{graphicx}
\usepackage{subfigure}
\usepackage{amssymb}
\usepackage{booktabs} 

\usepackage{amsmath}
 
\usepackage{hyperref}

\usepackage{color}

\title{Denoising Autoencoders for Overgeneralization in Neural Networks}

\author{
  Giacomo ~Spigler\thanks{ \texttt{http://www.spigler.net/giacomo} } \\
  The Biorobotics Institute\\
  Scuola Superiore Sant'Anna\\
  Pisa, Italy \\
  \texttt{giacomo.spigler@santannapisa.it} \\
}

\begin{document}

\maketitle

\begin{abstract}

Despite recent developments that allowed neural networks to achieve impressive performance on a variety of applications, these models are intrinsically affected by the problem of overgeneralization, due to their partitioning of the full input space into the fixed set of target classes used during training. Thus it is possible for novel inputs belonging to categories unknown during training or even completely unrecognizable to humans to fool the system into classifying them as one of the known classes, even with a high degree of confidence. This problem can lead to security problems in critical applications, and is closely linked to open set recognition and 1-class recognition. This paper presents a novel way to compute a confidence score using the reconstruction error of denoising autoencoders and shows how it can correctly identify the regions of the input space close to the training distribution.

\end{abstract}

\section{Introduction}


Discriminative models in machine learning, like neural networks, have achieved impressive performance in a variety of applications. Models in this class, however, suffer from the problem of overgeneralization, whereby the whole input space is partitioned between the set of target classes specified during training, and generally lack the possibility to reject a novel sample as not belonging to any of those.

\begin{figure}[!t]
\centering
\includegraphics[width=0.3\linewidth]{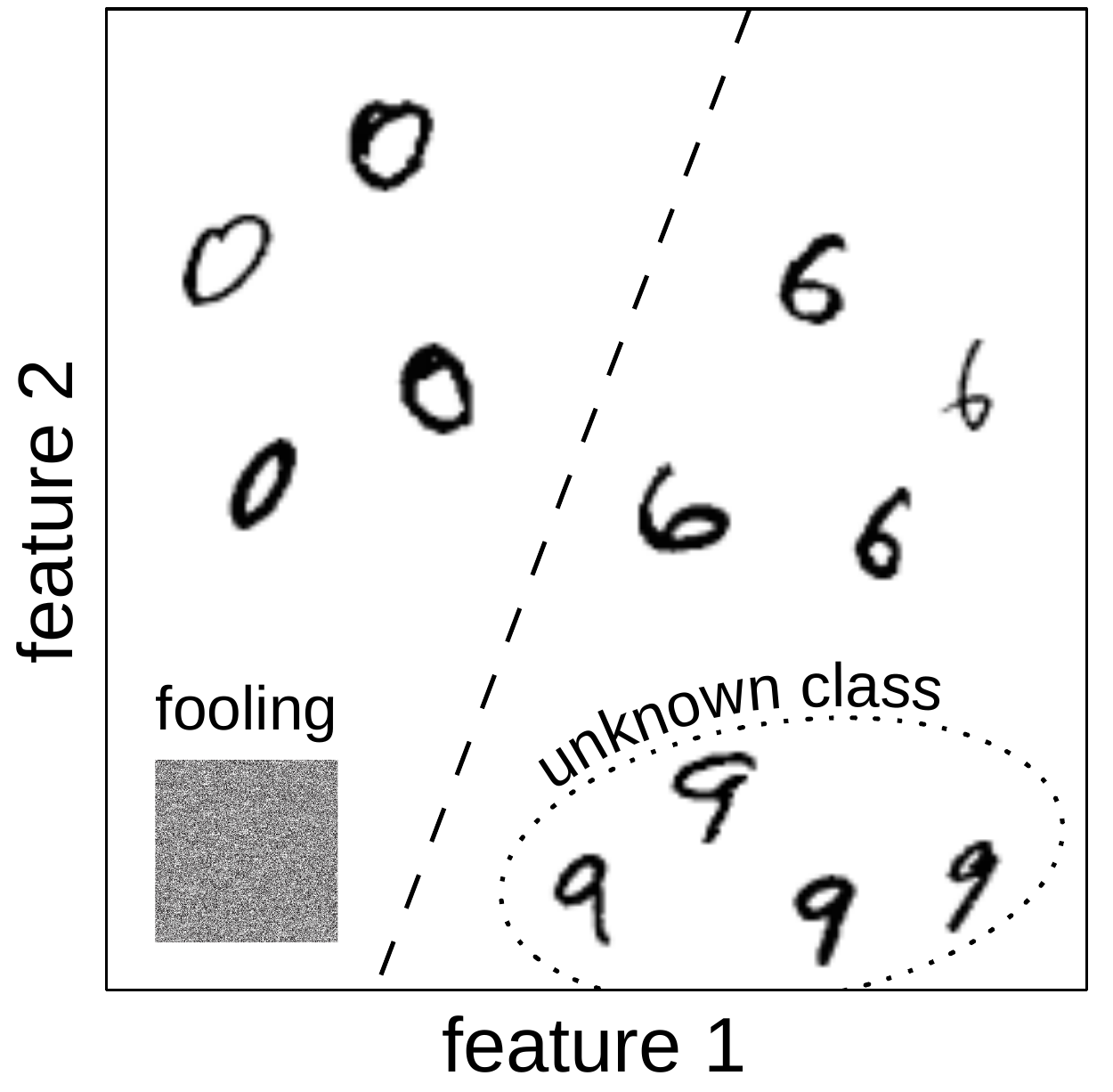}
\caption{ A linear classifier is trained to recognize exclusively pictures of digits `0' and `6'. Digit `9' was never observed during training, but in this example it is wrongly classified as digit `6'. This is an example of overgeneralization. A similar problem is `fooling', whereby it is possible to generate images that are unrecognizable to humans but are nonetheless classified as one of the known classes with high confidence, for example here the noise-looking picture in the bottom-left corner that is classified as digit `0'. }
\label{fig:overgeneralization_and_fooling}
\end{figure}

A main issue with overgeneralization is in the context of \emph{open set recognition} \cite{scheirer2013_opensetrecog_1vssetmachine} and \emph{open world recognition} \cite{bendale2015towards}, where only a limited number of classes is encountered during training while testing is performed on a larger set that includes a potentially very large number of unknown classes that have never been observed before. An example is shown in Figure \ref{fig:overgeneralization_and_fooling} where a linear classifier is trained to discriminate between handwritten digits `0' and `6'. As digit `9' is not present in the training set, it is here wrongly classified as `6'. In general, instances of classes that are not present in the training set will fall into one of the partitions of the input space learnt by the classifier. The problem becomes worse in real world applications where it may be extremely hard to know in advance all the possible categories that can be observed.

Further, the region of meaningful samples in the input space is usually small compared to the whole space. This can be easily grasped by randomly sampling a large number of points from the input space, for example images at a certain resolution, and observing that the chance of producing a recognizable sample is negligible. Yet, discriminative models may assign a high confidence score to such random images, depending on the learnt partition of the input space. This is indeed observed with \emph{fooling} \cite{fooling}, for which it was shown to be possible to generate input samples that are unrecognizable to humans but get classified as a specific target class with high confidence (see example in Figure \ref{fig:overgeneralization_and_fooling}). Fooling in particular may lead to security problems in critical applications.


As suggested in \cite{fooling}, these problems may be mitigated or solved by using generative models, that rather than learning the posterior of the class label $P(y|X)$ directly, learn the joint distribution $P(y,X)$ from which $P(X)$ can be computed. Modeling the distribution of the data would then give a model the capability to identify input samples as belonging to known classes, and to reject those that are believed to belong to unknown ones. Apart from mitigating the problem of overgeneralization, modeling the distribution of the data would also be useful for applications in novelty and outlier detection \cite{markou2003novelty} and incremental learning \cite{bendale2015towards}, broadening the range of applications the same model could be used in.

Estimating the marginal probability $P(X)$ is, however, not trivial. Luckily, computing the full distribution may not be necessary. The results in this work suggest that identification of high-density regions close to the local maxima of the data distribution is sufficient to correctly identify which samples belong to the distribution and which ones are to be rejected. Specifically, it is possible to identify and classify the critical points of the data distribution by exploiting recent work that has shown that in denoising \cite{denoising_autoencoder} and contractive \cite{contractive_autoencoder} autoencoders, the reconstruction error tends to approximate the gradient of the log-density. A measure of a confidence score can then be computed as a function of this gradient.


Here, a set of experiments is presented to compare the empirical performance of the proposed model with baselines and with the COOL (Competitive Overcomplete Output Layer) \cite{cool} model that has been recently applied to the problem of fooling.

\section{Overview of Previous Work}


The simplest way to limit overgeneralization in a given classifier is to set a threshold on the predicted outputs and rejecting any sample below its value (for example \cite{phillips2011evaluation, hendrycks2016baseline}). The output of the model is thus treated as an estimate of the confidence score of the classifier. This approach, however, was shown to be sensitive to the problem of fooling \cite{fooling}. Alternatively, a confidence score may be computed based on the k-Nearest-Neighbor algorithm (e.g., \cite{hautamaki2004outlier, zhao2009anomaly}).    


Another way to mitigate the problem, as done in classical object recognition, is to use a training set of positive samples complemented with a set of negative samples that includes instances belonging to a variety of `other' classes. This approach however does not completely solve the problem, and it is usually affected by an unbalanced training set due to the generally larger amount of negatives required. As the potential amount of negatives can be arbitrarily large, a further problem consists in gathering an amount of data sufficient to approximate their actual distribution, which is made even worse by the fact that the full set of negative categories may not be known when training the system. For example, in the context of object recognition in vision, high-resolution images may represent any possible image class, the majority of which is likely not known during training. The use of negative training instances may nonetheless mitigate the effect of categories that are known to be potentially observed by the system.

The problem of overgeneralization is further present in the context of `open set recognition', that was formally defined by Scheirer and colleagues \cite{scheirer2013_opensetrecog_1vssetmachine}. In this framework, it is assumed that a classifier is trained on a set of `known' classes and potentially on a set of `known unknown' ones (e.g., negative samples). Testing, however, is performed on a larger set of samples that include `unknown unknown' classes that are never seen during training. Models developed to address the problem of open set recognition focus on the problem of `unknown unknown' classes \cite{scheirer2014_wsvm}. The seminal paper that gave the first formal definition of the problem proposed the 1-vs-Set Machine algorithm as an extension to SVM that is designed to learn an envelope around the training data using two parallel hyperplanes, with the inner one separating the data from the origin, in feature space \cite{scheirer2013_opensetrecog_1vssetmachine}. Scheirer and colleagues then proposed the Weibull-calibrated SVM (W-SVM) algorithm to address multi-class open set recognition \cite{scheirer2014_wsvm}. Another interesting approach was recently applied to deep neural networks with the OpenMax model \cite{opensetdeepnetworks}, that works by modeling the class-specific distribution of the activation vectors in the top hidden layer of a neural network, and using the information to recognize outliers.


Related to the problem of open set recognition is that of `open world recognition', in which novel classes first have to be detected and then learnt incrementally  \cite{bendale2015towards}. This can be seen as an extension to open set recognition in which the `unknown unknown' classes are discovered over time, becoming `novel unknowns'. The new classes are then labelled, potentially in an unsupervised way, and become 'known'. The authors proposed the Nearest Non-Outlier (NNO) algorithm to address the problem.

A special case of open set recognition is 1-class recognition, in which training is performed on samples from a single class, with or without negative samples. The 1-Class SVM algorithm was proposed to address this problem \cite{oneclasssvm}, by fitting a hyperplane that separates all the data points from the origin, in feature space, maximizing its distance from the origin. The algorithm has been applied in novelty and outlier detection \cite{oneclasssvm_noveltydetection}. Variants of the algorithm like Support Vector Data Description (SVDD) have also been used to learn an envelope around points in the dataset \cite{oneclasssvm_support_vector_data_description_svdd}. Other systems have tried to estimate the boundaries of the data by computing the region of minimum volume in input space containing a certain probability mass \cite{park2010computable}. 

Finally, a specific sub-problem of overgeneralization is `fooling' \cite{fooling}. The ``Competitive Overcomplete Output Layer'' (COOL) model \cite{cool} was recently proposed to mitigate the problem of fooling. COOL works by replacing the final output layer of a neural network with a special COOL layer, constructed by replacing each output unit with $\omega$ ones (the degree of overcompleteness). The $\omega$ output units for each target class are then made to compete for activation by means of a softmax activation that forces them to learn to recognize different parts of the input space, overlapping only within the region of support of the data generating distribution. The network can then compute a confidence score as the product of the activation of all the units belonging to the same target class, that is high for inputs on which a large number of units agrees on, and low in regions far from the data distribution, where only few output units are active. 


\section{Proposed Solution}
\label{sec:methods}

The solution presented here is based on a novel measure of confidence in the correct identification of data points as belonging to the training distribution, or their rejection. Ideally, such a confidence score would be a function of the data probability $p(\mathbf{x})$. Computing the full distribution may however not be necessary. In particular, the problem can be simplified with the identification of points belonging to the data manifold as points that are close to the local maxima of the data generating distribution.


It has been recently shown that denoising \cite{denoising_autoencoder} and contractive \cite{contractive_autoencoder} autoencoders implicitly learn features of the underlying data distribution \cite{alain_bengio_2013, bengio_2013}, specifically that their reconstruction error approximates the gradient of its log-density

\begin{equation}
  \label{eqn:autoencoder_score}
  \frac{\partial \log p(\mathbf{x})}{\partial \mathbf{x}} \propto r(\mathbf{x})-\mathbf{x}
\end{equation}

for small corruption noise ($\sigma \rightarrow 0$). $r(\mathbf{x})=Dec(Enc(\mathbf{x}))$ is the reconstructed input. Larger noise is however found to work best in practice. The result has been proven to hold for any type of input (continuous or discrete), any noise process and any reconstruction loss, as long as it is compatible with a log-likelihood interpretation \cite{bengio_2013}. A similar interpretation suggested that the reconstruction error of regularized autoencoders can be used to define an energy surface that is trained to take small values on points belonging to the training distribution and higher values everywhere else \cite{ebgan}.

Thus, critical points of the data distribution correspond to points with small gradient of the log-density, that is small reconstruction error (Equation \ref{eqn:autoencoder_score}). Those are indeed points that the network can reconstruct well, and that it has thus hopefully experienced during training or has managed to generalize to well. A confidence score can thus be designed that takes high values for points on the data manifold, that is points near the local maxima of the log-density of the data distribution, and small values everywhere else. 

We note however that this approach cannot distinguish between local minima, maxima or saddle points (Figure \ref{fig:2dvisualization} shows such an example), and may thus assign a high confidence score to a small set of points not belonging to the target distribution. Here the problem is addressed by scaling the computed confidence by a function $\Gamma(\mathbf{x})$ that favours small or negative curvature of the log-density of the data distribution, which can in turn be computed from the diagonal of the Hessian, estimated from the Jacobian of the reconstruction function as shown in \cite{alain_bengio_2013}

\begin{equation}
  \label{eqn:hessian}
  \frac{\partial^2 \log p(\mathbf{x})}{\partial \mathbf{x}^2} \propto \frac{\partial r(\mathbf{x})}{\partial \mathbf{x}} - I
\end{equation}

A variety of functions may be defined with the desired characteristics, exploiting Equations \ref{eqn:autoencoder_score} and \ref{eqn:hessian}. One possible way, that we will use throughout this paper, is to compute the confidence score $\tilde{c}(\mathbf{x})$ as

\begin{equation}
  \label{eqn:pseudo_confidence}
  \tilde{c}(\mathbf{x}) = \exp\left(-\frac{\alpha}{D} \|r(\mathbf{x})-\mathbf{x}\|_2\right) \Gamma(\mathbf{x})
\end{equation}

\begin{align}
  \label{eqn:curvature_function}
  \Gamma(\mathbf{x}) =
    \begin{cases}
      1                                \hspace{2.3cm} \text{if } \gamma(\mathbf{x}) \leq 0 \\
      \exp(-\beta \gamma(\mathbf{x}))  \hspace{0.5cm} \text{if } \gamma(\mathbf{x})>0
    \end{cases}
\end{align}

\begin{equation}
  \label{eqn:curvature}
  \gamma(\mathbf{x}) = \frac{1}{D} \sum_i \left( \frac{\partial r_i(\mathbf{x})}{\partial x_i} - 1 \right) 
\end{equation}
where $D$ is the dimensionality of the inputs $\mathbf{x} = \left(x_1, x_2, \ldots, x_D \right)$, $\alpha$ a parameter that controls the sensitivity of the function to outliers and $\beta$ a parameter that controls the sensitivity to $\gamma(\mathbf{x})$, which is proportional to the average of the diagonal elements of the Hessian of the log-density at $x$ (from Equation \ref{eqn:hessian}). The first component of $\tilde{c}(\mathbf{x})$ identifies the extrema points of the log-density of the data (from Equation \ref{eqn:autoencoder_score}), while $\Gamma(\mathbf{x})$ is used to limit high values of the confidence scores to the maxima only (i.e., to points predicted to lie near the data manifold).


A classifier can finally be modified by scaling its predicted output probabilities $y$ by $\tilde{c}(\mathbf{x})$ computed using a denoising autoencoder trained together with the classifier
\begin{equation}
  \label{eqn:classifier_corrected_output}
  \tilde{\mathbf{y}} = \tilde{c}(\mathbf{x}) \mathbf{y}
\end{equation}

If the outputs of the classifier are normalized, for example using a softmax output, this can be interpreted as introducing an implicit `reject' option with probability $1-\tilde{c}(\mathbf{x})$. The confidence score proposed here, however, was not designed as a probability estimate.

In the experiments presented here, the classifier is constructed as a fully connected softmax layer attached on top of the top hidden layer of an autoencoder with symmetric weights (i.e., attached to the output of the encoder), in order to keep the number of weights similar (minus the bias terms of the decoder) to an equivalent feed-forward benchmark model, identical except for its lack of the decoder. In general, keeping the autoencoder separate from the classifier or connecting the two in more complex ways will work, too, as well as using a classifier that is not a neural network. In case the autoencoder and the classifier are kept separate, the autoencoder is only used to infer information about the data distribution. Pairing the systems together, however, might provide advantages outside the scope of the present work, like enabling a degree of semi-supervised learning. The autoencoder may also be further improved by replacing it with the discriminator of an EBGAN \cite{ebgan} to potentially learn a better model of the data.


\section{Experiments}

\subsection{2D example}

\begin{figure}[!t]
\centering
\includegraphics[width=0.4\linewidth]{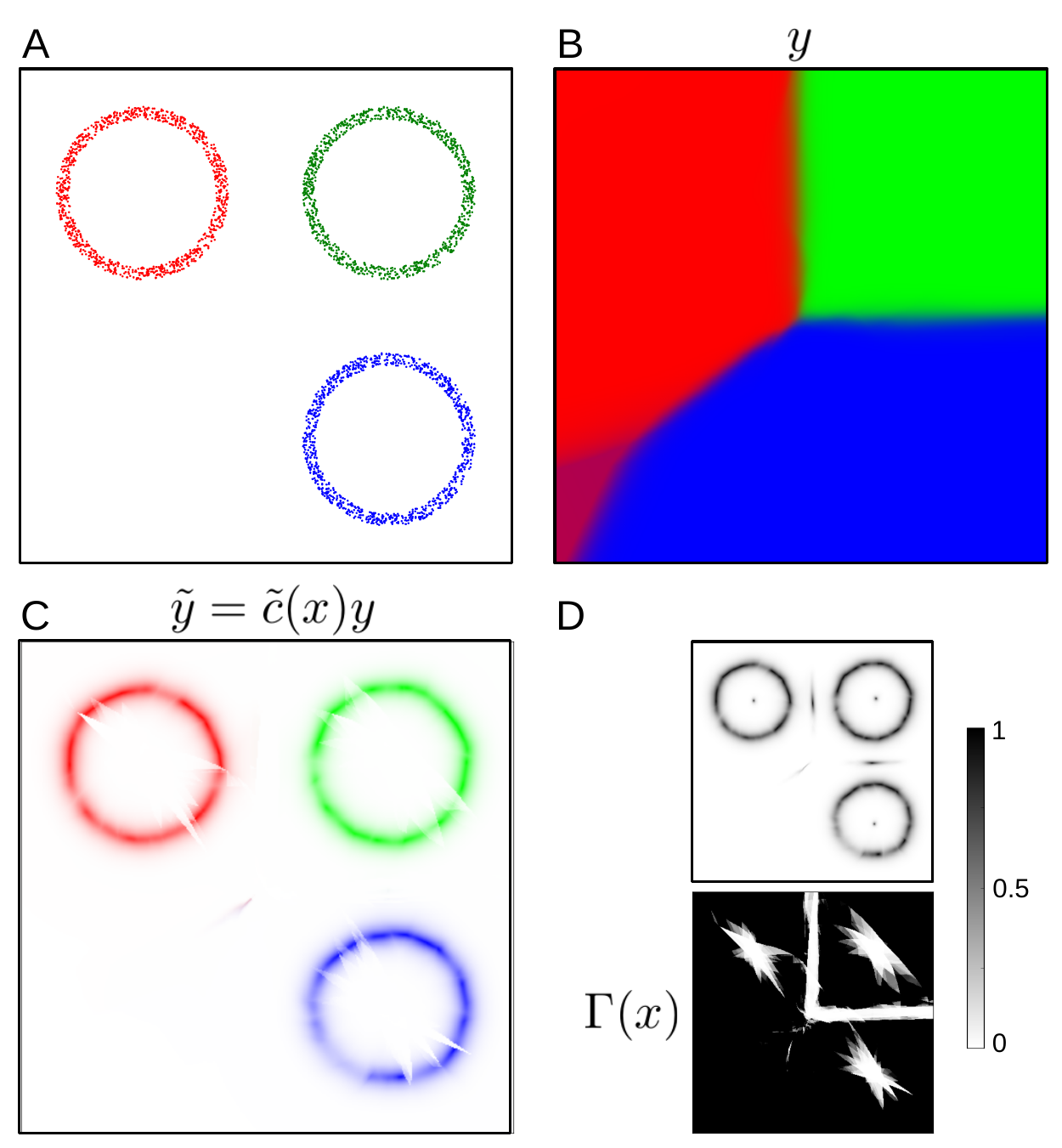}
\caption{ The system presented here is trained to classify points sampled from three uniform ring distributions. \textbf{A.} $1000$ data points are sampled from each of the target distributions. \textbf{B.} Labeling $y$ of each point in the input space without scaling of the classifier's output by the confidence score. \textbf{C.} Labeling $\tilde{\mathbf{y}}$ of each point in the input space scaled by the computed confidence score. Regions in white are assigned low confidence scores.  \textbf{D.} Top: confidence score without $\Gamma(\mathbf{x})$. Bottom: estimate of the curvature of the log-distribution of the data ($\Gamma(\mathbf{x})$). The confidence score $\tilde{c}(\mathbf{x})$ is the product of the two functions. The panel in \textbf{B} is the product of the classifier's output (\textbf{C}) and the confidence score. }
\label{fig:2dvisualization}
\end{figure}

The model was first tested on a 2D classification task to visualize its capacity to learn the support region of the input space of each training class. Three target distributions were defined as uniform rings with thickness of $0.1$, inner radius of $0.6$ and centers $(-1, 1)$, $(1, 1)$ and $(1, -1)$. The training distributions are shown in Figure \ref{fig:2dvisualization}A. Training was performed with minibatches of size $64$ using the Adam optimizer \cite{adam_optimizer} for a total of $50000$ update steps. As shown in Figure \ref{fig:2dvisualization}B, the model learned to correctly identify the support region of the target distributions. On the contrary, the uncorrected classifier partitioned the whole space into three regions, incorrectly labeling most points (panel C). The confidence score computed by the model presented here (panel D) helps the system to prevent overgeneralization by limiting the decisions of the classifier to points likely to belong to one of the target distributions. Panel D further evidences the different contributions of the two factors used to compute the confidence score. A measure of proximity to any local extrema of the data generating distribution (top part of panel D) is modulated to remove the local minima using the $\Gamma(\mathbf{x})$ function of the local curvature of the log probability of the distribution (bottom part of panel D). It is important to observe, however, that the $\Gamma(\mathbf{x})$ function may reduce the computed confidence score of valid samples. Different types of applications may benefit from its inclusion or exclusion, depending on whether more importance is given to the correct rejection of samples that do not belong to the training distribution or to their correct classification, to the expense of a partial increase in overgeneralization.

\subsection{Fooling}
\label{sec:fooling}

The model presented in this paper was next tested on a benchmark of fooling on the MNIST \cite{mnist} and Fashion-MNIST \cite{fmnist} datasets similar to the one proposed in \cite{cool}. However, contrary to the previous work, the classification accuracy of the models is reported as a `thresholded classification accuracy', that is by additionally requiring test samples to be assigned a confidence score higher than the threshold used to consider a fooling instance valid. This metric should indeed be reported alongside the fooling rate for each model, as otherwise a model that trivially limits the confidence scores of a network to a fixed value lower than the threshold used to consider fooling attempts to be valid would by definition never be fooled. The same model would however never classify any valid sample above that same threshold. This metric thus proves useful to compare different models with varying degrees of sensitivity to overgeneralization.  

The fooling test was performed by trying to fool a target network to classify an input that is unrecognizable to humans into each target class (digits $0$ to $9$). The fooling instances were generated using a Fooling Generator Network (FGN) consisting of a single layer perceptron with sigmoid activation and an equal number of input and output units (size of $(28,28)$ here). Most importantly, the FGN produces samples with values bounded in $(0,1)$ without requiring an explicit constraint. Fooling of each target digit was attempted by performing stochastic gradient descent on the parameters of the FGN to minimize the cross-entropy between the output of the network to be fooled and the specific desired target output class. Fooling of each digit was attempted for $20$ trials with different random inputs to the FGN, each trial consisting of up to $10000$ parameter updates, as described in \cite{cool}.

In the first test we compared three models, a plain Convolutional Neural Network (CNN), the same CNN with a Competitive Overcomplete Output Layer (COOL) \cite{cool}, and a network based on the system described in Section \ref{sec:methods}, built on the same CNN as the other two models with the addition of a decoder taking the activation of the top hidden layer of the CNN as input, to compute the dAE-based confidence score $\tilde{c}(\mathbf{x})$. The denoising autoencoder (dAE) was trained with corruption of the inputs by additive Gaussian noise. All the models were trained for a fixed $100$ epochs. Fooling was attempted at two different thresholds, $90\%$ and $99\%$, in contrast to the previous work that used only the $99\%$ one \cite{cool}. Comparing the models at different thresholds indeed gives more information about their robustness and may amplify their differences, thus improving the comparison. Tables \ref{table:results1_fooling_mnist} and \ref{table:results1_fooling_fmnist} reports the results for the three models, with the further splitting of the denoising autoencoder model in two separate tests, using either a separate decoder (\emph{dAE asym}) or building the decoder as a symmetric transpose of the encoder (with the addition of new bias parameters; \emph{dAE sym}). The table reports the thresholded classification accuracy for all the models together with the original, unthresholded one. Fooling was measured as the proportion of trials ($200$ total, $20$ repetitions of $10$ digits) that produced valid fooling samples within the maximum number of updates. The average number of updates required to fool each network is reported in parentheses. The full set of parameters used in the simulations is reported in Appendix \ref{appendix:foolingmnist}. The model presented here outperformed the other two at both thresholds, while also retaining a high thresholded classification accuracy, even at high thresholds. As in the previous protocol \cite{cool}, the cross-entropy loss used to optimize the FGN was computed using the unscaled output $y$ of the network.

\begin{table}[!t]
\renewcommand{\arraystretch}{1.5}
\scriptsize
\caption{MNIST fooling results}
\label{table:results1_fooling_mnist}
\centering
\hspace{-0.1in}
\begin{tabular}{|c| c c c | c c |}
\hline
\multicolumn{1}{|c|}{Model} & \multicolumn{3}{c|}{Accuracy} & \multicolumn{2}{c|}{Fooling Rate (Avg Steps)} \\
\multicolumn{1}{|c|}{} & \multicolumn{1}{c}{0\%} & \multicolumn{1}{c}{$90\%$} & \multicolumn{1}{c|}{$99\%$} & \multicolumn{1}{c}{$90\%$} & \multicolumn{1}{c|}{$99\%$} \\ \hline

CNN &          99.35\% & 99.23\% & 99\% &    100\% (63.5) & 99\% (187.1) \\ \hline
COOL &         99.33\%  & 98.1\% & 93.54\% &    34.5\% (238.8) & 4.5\% (313.4) \\ \hline
dAE sym &   98.98\% & 98.11\% & 96.8\%     &    \textbf{0\% (-)} & \textbf{0\% (-)} \\ \hline
dAE asym &  99.14\% & 98.41\% &            &   \textbf{0\% (-)} &         \\ \hline

\end{tabular}
\end{table}

\begin{table}[t!]
\renewcommand{\arraystretch}{1.5}
\scriptsize
\caption{Fashion-MNIST fooling results}
\label{table:results1_fooling_fmnist}
\centering
\hspace{-0.1in}

    \begin{tabular}{|c| c c c | c c |}
      \hline
        \multicolumn{1}{|c|}{Model} & \multicolumn{3}{c|}{Accuracy} & \multicolumn{2}{c|}{Fooling Rate (Avg Steps)} \\
        \multicolumn{1}{|c|}{} & \multicolumn{1}{c}{0\%} & \multicolumn{1}{c}{$90\%$} & \multicolumn{1}{c|}{$99\%$} & \multicolumn{1}{c}{$90\%$} & \multicolumn{1}{c|}{$99\%$} \\ \hline \hline

        CNN &          91.65\% & 90.91\% & 89.27\% &    100\% (113.0) & 30.5\% (902.0) \\ \hline
        COOL &          91.23\% & 87\% & 65.3\% &    \textbf{0\% (-)} & \textbf{0\% (-)} \\ \hline
        dAE sym &          91.59\% & 77.8\% & 64.87\% &    \textbf{0\% (-)} & \textbf{0\% (-)} \\ \hline

  \end{tabular}
\end{table}

As the difference between the autoencoders using a symmetric versus asymmetric decoder was found to be minimal on MNIST, the symmetric autoencoder was used for all the remaining experiments, so that the three models had a similar number of parameters ($1.31M$ for CNN and dAE, $1.35M$ for COOL).

We further observed that the results in Table \ref{table:results1_fooling_mnist} were different from those reported in \cite{cool}. Specifically, the fooling rate of the COOL was found to be significantly lower than that reported ($47\%$), as well as the average number of updates required to fool it (more than $5000$). The major contributor to this difference was found to be the use of Rectified Linear Units (ReLUs) in the experiments reported here, compared to sigmoid units in the original study. This was shown in a separate set of simulations where all the three models used sigmoid activations instead of ReLUs and a fixed fooling threshold of $99\%$. In this case the thresholded classification accuracy of the models was slightly higher ($98.39\%$ for the plain CNN, $96.55\%$ for COOL, and $96.58\%$ for dAE), but it was matched with a significant increase in the fooling rate of the COOL model ($95.5\% (2203.9)$; plain CNN $91\% (519.2)$, dAE $0\%$). Other variations in the protocol that could further account for the differences found could be the different paradigm for training ($100$ fixed training epochs versus early stopping on a maximum of $200$ epochs) and a slightly different network architecture, that in the present work used a higher number of filters at each convolutional layer.

Next, the effect of the learning rate used in the fooling update steps was investigated by increasing it from the one used in the previous study ($\eta=0.00001$) to the one used to train the models $\eta=0.001$, expecting a higher fooling rate. The threshold was set to $90\%$. Indeed, the plain CNN was found to be fooled on $100\%$ of the trials in just $2.66$ updates, while the dAE based model was still never fooled. COOL, on the other hand, significantly decreased in performance, with a fooling rate of $56.5\%$ ($878.3$ average updates).

Finally, the COOL and dAE models were tested by attempting to fool their confidence scores directly, rather than their output classification scores, in contrast to \cite{cool} (i.e., using $\tilde{\mathbf{y}}$ instead of $\mathbf{y}$ for the cross-entropy loss used to update the FGN). A threshold of $99\%$ was used.  Interestingly, the COOL model was never fooled, while the model described here was fooled on $1\%$ of the trials, although requiring a large number of updates ($5470.8$ on average). Also, it was found that while adding $L_2$ regularization to the weights of the dAE model led to a significantly higher fooling rate ($100\%$ rate in $6500.3$ average updates for $\lambda_{L_2}=10$), the generated samples actually resembled real digits closely, and could thus not be considered examples of fooling. This shows that the dAE model, when heavily regularized, is capable of learning a tight boundary around the high density regions of the data generating distribution, although at the cost of reducing its thresholded accuracy ($87.84\%$ for $\lambda_{L_2}=10$). The set of generated samples is shown as Supplementary Figure D for $\lambda_{L_2}=\{10,100\}$.

\begin{figure}[!t]
\centering
\includegraphics[width=0.6\linewidth]{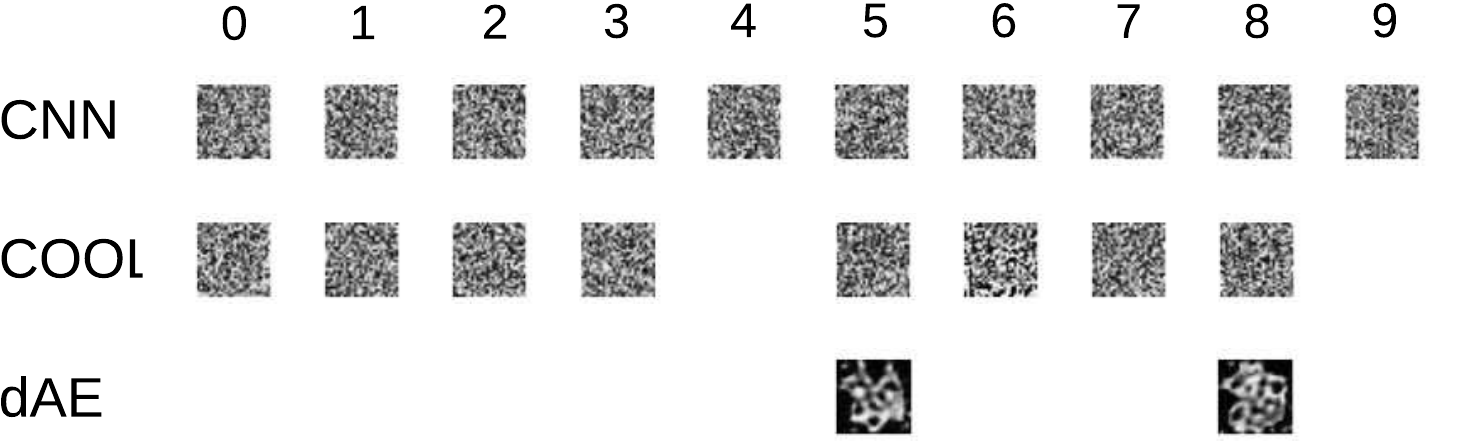}
\caption{ Visualization of a set of generated fooling samples from the main results of Table \ref{table:results1_fooling_mnist}. The samples from the plain CNN and the COOL models were computed by trying to fool each system's output classification scores above a threshold of $90\%$. As fooling was unsuccessful on the dAE model in this case, the results reported here were taken from the simulations in which fooling was performed directly on the output scaled by the confidence score ($\tilde{\mathbf{y}}$). }
\label{fig:fooling_samples}
\end{figure}

An example of the generated fooling samples is reported in Figure \ref{fig:fooling_samples}, showing instances from the main results of table \ref{table:results1_fooling_mnist} for the plain CNN and COOL, and for the experiment with fooling the confidence scores directly for the dAE model.

\subsection{Open Set Recognition}

The three models that were tested on fooling, a plain CNN, COOL \cite{cool} and the dAE model described in this paper were next compared in the context of open set recognition.

Open set recognition was tested by building a set of classification problems with varying degrees of openness based on the MNIST and Fashion-MNIST datasets. Each problem consisted in training a target model only on a limited number of `known` training classes (digits) and then testing it on the full test set of $10$ digits, requiring the model to be able to reject samples hypothesized to belong to `unknown' classes.
The degree of openness of each problem was computed similarly to \cite{scheirer2013_opensetrecog_1vssetmachine}, as
$$
  openness = 1 - \sqrt{\frac{num\_training\_classes}{num\_total\_classes}}
$$
where $num\_training\_classes$ is the number of `known' classes seen during training and $num\_total\_classes$ is $10$ for both datasets. A high value of openness reflects a larger number of unknown classes seen during testing than that of classes experienced during training. The number of training classes was varied from $1$ to $10$, reflecting the full range of degrees of openness offered by the dataset.

For each fixed number of training classes used in training, the models were trained for $10$ repetitions on different random subsets of the digits, to balance between easier and harder problems depending on the specific digits used. The same subsets of digits were used for all the three models. Correct classification was computed as a correct identification of the class label and a confidence score above a classification threshold of $99\%$, while correct rejection was measured as either assigning a low classification score (below $99\%$) or classifying the input sample as any of the classes not seen during training (for simplicity, the networks used a fixed number of output units for all the problems, with the target outputs corresponding to the `unknown' classes always set to zero). The models were trained for a fixed $100$ epochs for each task.

\begin{figure}[!t]
\centering
\includegraphics[width=0.8\linewidth]{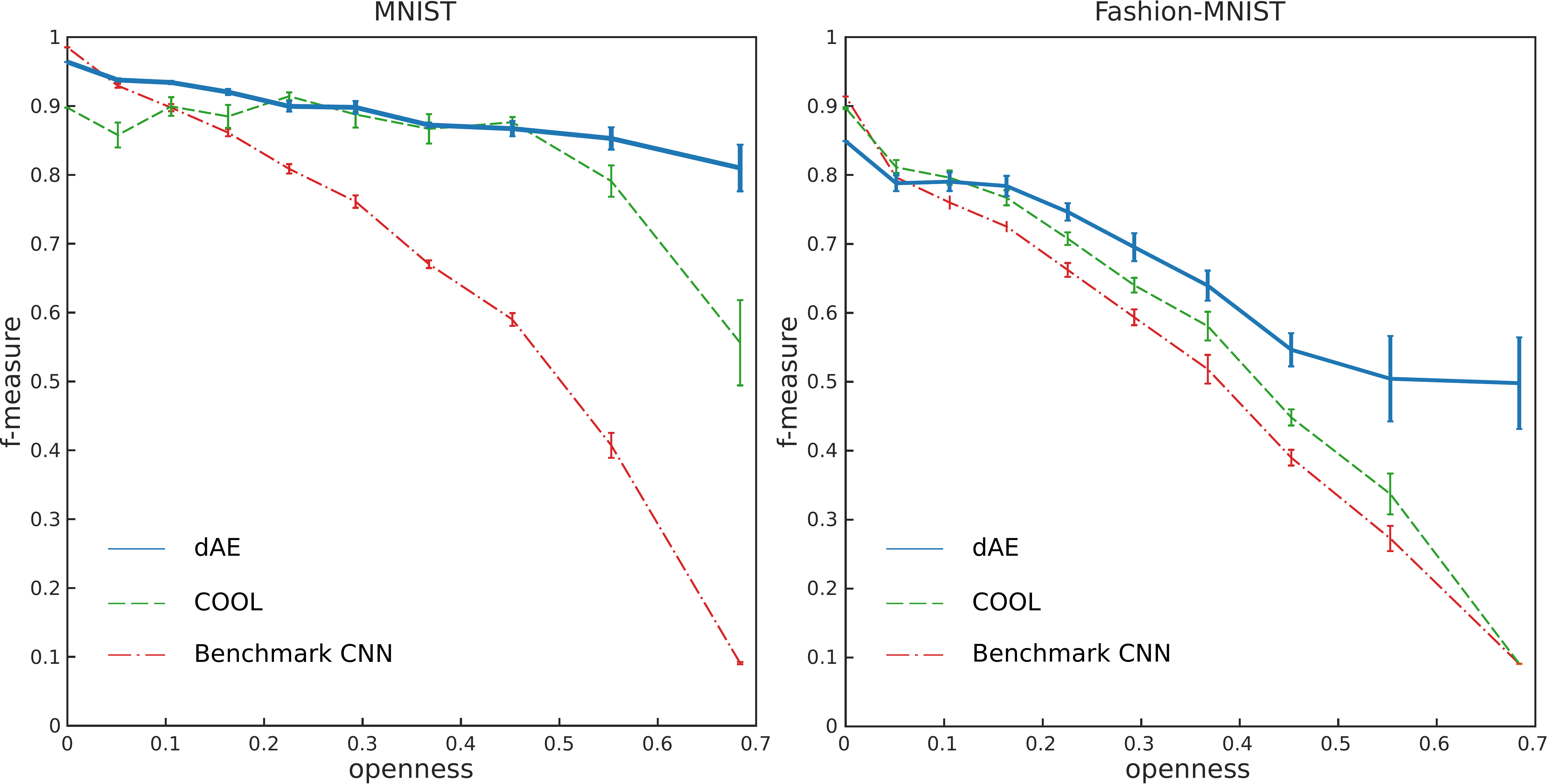} 
\caption{ Comparison of the three models on a benchmark of open set recognition. The F-measure was computed for each model on problems created from the MNIST (left) and Fashion-MNIST (right) datasets by only using a limited number of `known' classes during training while testing on the full test set (e.g., training on classes $0$ and $3$ but testing on all classes $[0,9]$), requiring the models to be able to reject samples belonging to `unknown' classes. Higher values for the openness of a problem reflect a smaller number of classes used during training. The curves are averaged across $10$ runs using different sub-sets of digits. Error bars denote standard deviation.  } 
\label{fig:opensetrecognition_99pct}
\end{figure}

Figure \ref{fig:opensetrecognition_99pct} reports the results of the experiment. Like in the previous published benchmarks on open set recognition \cite{scheirer2013_opensetrecog_1vssetmachine, scheirer2014_wsvm, opensetdeepnetworks}, the performance of the models for each degree of openness (indexed by $i$) was computed as the F-measure, the harmonic mean of the precision and recall scores, averaged across all the repetitions for the same degree of openness.

$$
  F_i = 2 \times \frac{precision_i \times recall_i}{precision_i + recall_i}
$$

Results from a similar experiment with a lower threshold of $90\%$ are available as Supplementary Figure F.

\subsection{1-Class Recognition}
\label{sec:oneclassrecog}

The limit of open set recognition in which a single training class is observed during training, that is the problem of 1-class 
recognition, was next explored, comparing the model presented in this paper with COOL \cite{cool} and 1-Class SVM \cite{oneclasssvm}.

A separate 1-class recognition problem was created from the MNIST and Fashion-MNIST datasets for each target class. For each problem the models were trained using only samples from the selected class, while they were tested on the full test set of $10$ digits. No negative samples were used during training. Each model was trained for $100$ epochs on each problem.

\begin{figure}[!t]
\centering
\includegraphics[width=0.8\linewidth]{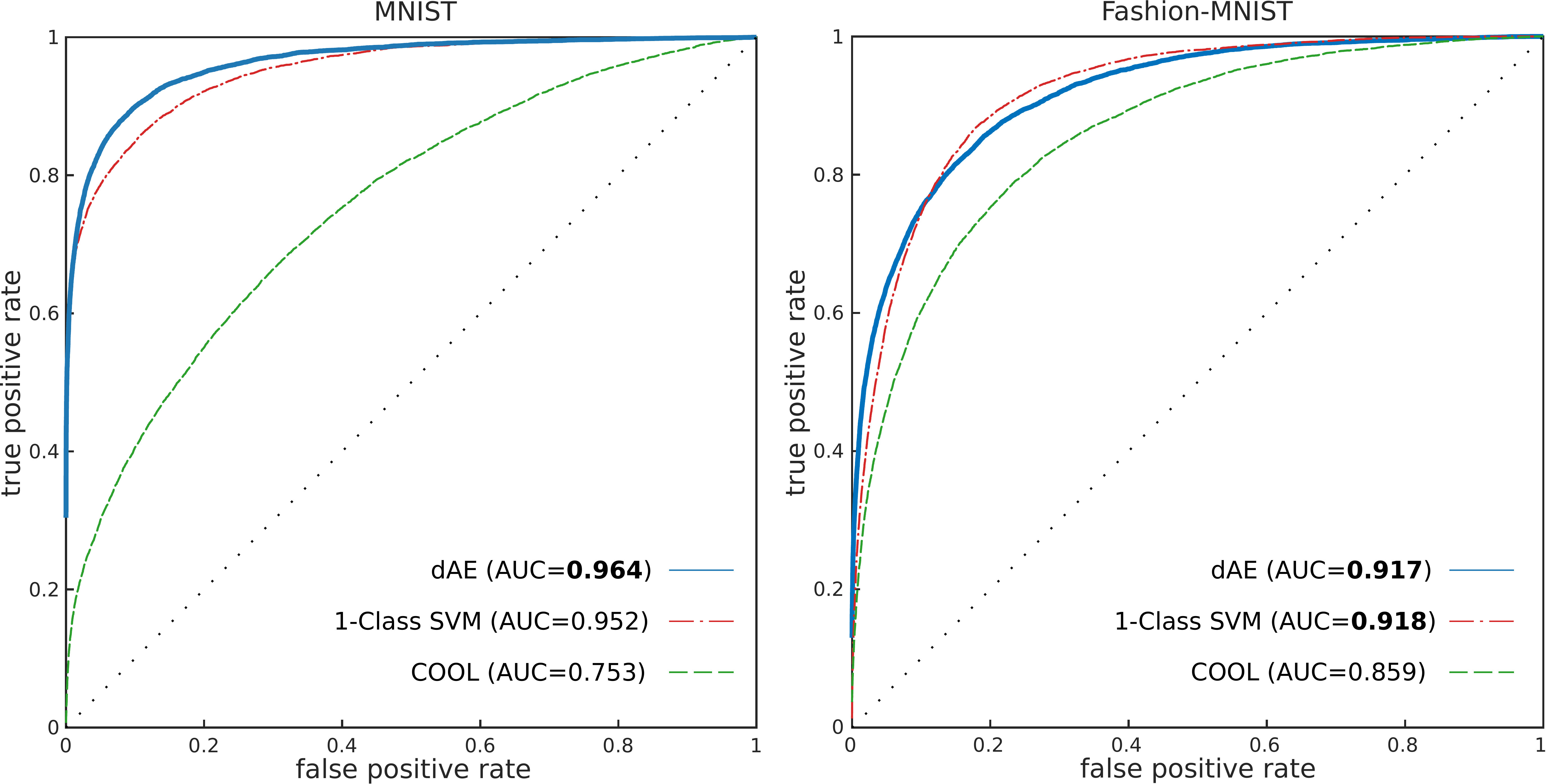}
\caption{ ROC curves averaged over $10$ 1-class recognition problems, one for each class in MNIST (left) and Fashion-MNIST (right), for three models, the dAE model described in this paper, 1-Class SVM \cite{oneclasssvm} and COOL \cite{cool}.  }
\label{fig:oneclass_recog_results}
\end{figure}

Figure \ref{fig:oneclass_recog_results} shows the results as a ROC curve averaged over the curves computed for each of the $10$ 1-class recognition problems. The dAE based model outperforms the other two, with an Area Under the Curve ($AUC$) of $0.964$, compared to $AUC=0.952$ of 1-Class SVM and $AUC=0.753$ of COOL.

\section{Discussion}

The confidence score that was introduced in this paper was found to perform better than a set of competing models in open set recognition and 1-class recognition. The system was also found to be significantly more robust to the problem of fooling than the state of the art COOL model. Together, these results show that it is possible to use information about the data generating distribution implicitly learnt by denoising autoencoders in meaningful ways, even without explicitly modeling the full distribution.

It is to be noted that when comparing the results to the COOL model we used the same degree of overcompleteness ($\omega=10$) as in the original paper. However, fine tuning of the parameter and in particular using higher values may achieve higher performance on the benchmarking tasks used here. Also, similarly to the original COOL paper, fooling was attempted on the output of the classifier, rather than directly on the confidence scores. This gives an advantage to systems in which the confidence score is computed in more complex ways, not directly dependendent on the output of the classifier. However, further tests as presented in Section \ref{sec:fooling} showed that the system presented here significantly outperforms the other models even when fooling is attempted directly on the confidence scores. In this particular case, it was further found that training the denoising autoencoder with heavy regularization resulted in generated samples resembling real digits, thus showing that the model had learnt a tight boundary around the data manifold.




It is interesting that the Energy-Based GAN (EBGAN) \cite{ebgan} makes use of the reconstruction error of a denoising autoencoder in a way compatible with the interpretation proposed here. In particular, it uses it as an approximated energy function that is learnt by the autoencoder to take low values for points belonging to the training distribution and high values everywhere else. As we have seen in Equation \ref{eqn:autoencoder_score}, it has been shown that the reconstruction error of denoising autoencoders is proportional to the gradient of the log-density of the data. Thus, small absolute values of the reconstruction error correspond to extrema points of the distribution, not limited to local maxima but also including minima and saddle points. If Figure \ref{fig:2dvisualization} were a good example of the dynamics of the system even on more complex data, then the problem of local minima and saddle points may be limited. However, if that was not the case, then EBGAN might learn to generate samples from regions of local minima of the data distribution, which may not be desirable. It would be interesting to modify the system using the $\Gamma(\mathbf{x})$ function described here (Equation \ref{eqn:curvature_function}) in order to correctly isolate only the local maxima of the distribution.

It would also be interesting to apply the regularization function used in EBGAN to the present model, adding a Pulling-away Term (PT) that forces learnt representations to be maximally different for different data points, by attempting to orthogonalize each pair of samples in a minibatch \cite{ebgan}. The stronger regularization may help the denoising autoencoder to learn a better representation of the data manifold, thus improving the confidence score $\tilde{c}(\mathbf{x})$.

Further improvements in the performance of the system may be achieved by separating the classifier and the denoising autoencoder, although combining the two may have other advantages, like adding a degree of semi-supervised learning or regularization of the autoencoder. It may also be possible to train an autoencoder to reconstruct hidden representations produced by pre-trained models, thus relying on more stable feature vectors rather than high-dimensional raw inputs. 

\section{Conclusion}

This paper presented a novel approach to address the problem of overgeneralization in neural networks by pairing a classifier with a denoising or contractive autoencoder that is used to compute a confidence score that assigns high values only for input vectors likely to belong to the training data distribution. In particular, recognition of an input as belonging to the distribution is performed by using an approximation of the gradient of the log-density and its curvature at the specific input point, and using this information to determine whether it lies close to a local maximum of the distribution. 


We have further explored the application of the system in the context of open set recognition. In general, the model presented here could be used in more complex architectures to allow for incremental and continual learning, by learning to recognize the regions of input space that have already been explored and learnt and potentially provide for different training regimes in the unexplored parts, in case new samples from those regions were to be observed in the future. For example, it may be applied to a system to allow for adding novel target classes even after deployment, without requiring a full re-training that may be costly in terms of compute time required, especially for large models. Similar to open set recognition is also 1-class recognition, that has proven to be a challenging problem. Building systems capable of robust 1-class recognition has critical applications in the detection of novelty, outliers and anomalies.

In conclusion, developing discriminative models capable of capturing aspects of the data distribution, even without explicitly modeling it, can prove very useful in a large number of practical applications, and future work on the topic will be highly beneficial. Here a system was presented to address the problem and was shown to perform better than other previously proposed systems on a set of benchmarks.


%


\section*{APPENDIX}

\section*{Details of the simulations}
\renewcommand{\thesubsection}{\alph{subsection}}

The models were trained on a cross-entropy loss by Stochastic Gradient Descent using the ADAM algorithm \cite{adam_optimizer} with $\eta=0.001$, $\beta_1=0.9$ and $\beta_2=0.999$. Tensorflow \cite{tensorflow} was used for the experiments. 

\subsection{2D example}

The dAE model used parameters $\alpha=40$, $\beta=5$ and $\sigma=0.2$, and a symmetric denoising autoencoder with inputs of size $2$ and two hidden layers both of size $200$. The classifier was a fully-connected layer attached to the top hidden layer of the autoencoder and had $3$ output units. Training was performed for $50000$ steps with minibatches of size $64$. The three target distributions were defined as uniform rings with thickness of $0.1$ and inner radius of $0.6$, centered at the three points $(-1, 1)$, $(1, 1)$ and $(1, -1)$.

\subsection{Fooling}
\label{appendix:foolingmnist}

The models compared are a regular CNN, the same CNN with the output layer replaced with a COOL layer (degree of overcompleteness $\omega=10$, as in \cite{cool}), and the same CNN with the addition of a decoder connected to the top hidden layer of the CNN, to complete the denoising autoencoder used to compute the confidence score $\tilde{c}(\mathbf{x})$. The architecture of the CNN is $\{Conv2D(1 \rightarrow 32, 5 \times 5),~ MaxPool(2 \times 2),~ Conv2D(32 \rightarrow 64, 5 \times 5),~ MaxPool(2 \times 2),~ FullyConnected(64 \rightarrow 400),~ FullyConnected(400 \rightarrow 10)\}$. Each layer is followed by a ReLU non-linearity, except for the output layer that is followed by a softmax. Fooling was attempted for $20$ times for each digit, each for up to $10000$ update steps with a learning rate for updating the FGN set to $\eta=0.00001$ as in \cite{cool}. Training was performed for $100$ epochs for each model. The dAE model was trained with additive Gaussian noise with zero mean and $\sigma=0.2$ for MNIST, $\sigma=0.1$ for Fashion-MNIST, and parameters $\beta=10$ and $\alpha$ variable depending on the threshold used ($\alpha=20$ for the $90\%$ classification threshold, $\alpha=3$ for the $99\%$ threshold on MNIST, and $\alpha=2$ for the $99\%$ threshold on Fashion-MNIST). All models trained on the Fashion-MNIST dataset used $L_2$ regularization with $\lambda_{L_2}=10$ (i.e., CNN, COOL and dAE).


\subsection{Open Set Recognition}

The Open Set Recognition tests used the same models as for the MNIST fooling ones, with a single threshold of $99\%$.


\subsection{1-Class Recognition}

The COOL and dAE models used the same parameters as the other experiments, except for the MNIST experiments in which $L_2$ regularization of the weights was used ($\lambda_{L_2}=10$) for the dAE model, as well as $\sigma=0.3$. 1-Class SVM was trained using the scikit-learn library \cite{scikit-learn}, and used $\nu=0.1$ and an RBF kernel ($\gamma=0.1$).


\bibliographystyle{unsrt}
\bibliography{references.bib}
%


\end{document}